\documentclass{tlp}
\usepackage{mathptmx}
\usepackage{amsmath}
\usepackage{algorithm,algpseudocode}
\usepackage{graphicx}
\usepackage{fancyvrb}
\usepackage{times}
\usepackage{soul}
\usepackage{url}
\usepackage{booktabs}
\usepackage{multirow}
\usepackage{xcolor}
\definecolor{linkcolor}{RGB}{52,59,144}
\usepackage[hidelinks, colorlinks=true, allcolors=linkcolor]{hyperref}

\newtheorem{exmp}{Example}
\algnewcommand\algorithmicinput{\textbf{Input:}}
\algnewcommand\Input{\item[\algorithmicinput]}
\algnewcommand\algorithmicoutput{\textbf{Output:}}
\algnewcommand\Output{\item[\algorithmicoutput]}
\algnewcommand\algorithmicforeach{\textbf{for each}}
\algnewcommand{\LeftComment}[1]{\Statex \(\triangleright\) #1}
\algdef{S}[FOR]{ForEach}[1]{\algorithmicforeach\ #1\ \algorithmicdo}

\algdef{S}[FOR]{ForEach}[1]{\algorithmicforeach\ #1\ \algorithmicdo}

\begin{document}
\lefttitle{Cambridge Author}

\jnlPage{1}{8}
\jnlDoiYr{2021}
\doival{10.1017/xxxxx}

  \title[FOLD-RM]
        {FOLD-RM: A Scalable, Efficient, and Explainable Inductive Learning Algorithm for Multi-Category Classification of Mixed Data
        %\thanks{Authors acknowledge support from NSF grants IIS 1718945, IIS 1910131, IIP 1916206 and US DoD.}  
        }

\begin{authgrp}
\author{Huaduo Wang, Farhad Shakerin, Gopal Gupta}
\affiliation{The University of Texas at Dallas, Richardson, USA}
\tt {$\{$huaduo.wang, farhad.shakerin, gupta$\}$@utdallas.edu}
% \email{$\{$huaduo.wang, farhad.shakerin, gupta$\}$@utdallas.edu}
\end{authgrp}

%   \author[H. Wang, F. Shakerin and G. Gupta]
%          {HUADUO WANG, FARHAD SHAKERIN, GOPAL GUPTA\\
%          The University of Texas at Dallas, Texas, USA\\
%         %  \email{$\{$huaduo.wang, farhad.shakerin, gupta$\}$@utdallas.edu}
%          \email{$\{$huaduo.wang, farhad.shakerin, gupta$\}$@utdallas.edu}
%          }

\newtheorem{lemma}{Lemma}[section]

\nocite{*}% includes all entries of BibTeX database into the list of references.

\maketitle

%\section*{Acknowledgment}

%
% The "title" command has an optional parameter, allowing the author to define a "short title" to be used in page headers.
\title{FOLD-RM}

%
% The "author" command and its associated commands are used to define the authors and their affiliations.
% Of note is the shared affiliation of the first two authors, and the "authornote" and "authornotemark" commands
% used to denote shared contribution to the research.

%
% By default, the full list of authors will be used in the page headers. Often, this list is too long, and will overlap
% other information printed in the page headers. This command allows the author to define a more concise list
% of authors' names for this purpose.

%
% The abstract is a short summary of the work to be presented in the article.
\begin{abstract}
FOLD-RM is an automated inductive learning algorithm for learning default rules for mixed (numerical and categorical) data. It generates an (explainable) answer set programming (ASP) rule set for \textit{multi-category classification} tasks while maintaining efficiency and scalability. The FOLD-RM algorithm is competitive in performance with the widely-used, state-of-the-art algorithms such as XGBoost and multi-layer perceptrons (MLPs), however, unlike these algorithms, the FOLD-RM algorithm produces an explainable model. FOLD-RM outperforms XGBoost on some datasets, particularly large ones. FOLD-RM also provides human-friendly explanations for predictions.
\end{abstract}
%GG412: abstract updated slightly


%
% Keywords. The author(s) should pick words that accurately describe the work being
% presented. Separate the keywords with commas.
% hw 03/31
\keywords{ Explainable AI, Data Mining, Inductive Logic Programming, Machine Learning}

%
% A "teaser" image appears between the author and affiliation information and the body 
% of the document, and typically spans the page. 

%
% This command processes the author and affiliation and title information and builds
% the first part of the formatted document.

\section{Introduction}
Dramatic success of machine learning has led to an avalanche of applications of Artificial Intelligence (AI). However, the effectiveness of these systems is limited by the machines' current inability to explain their decisions to human users. That is mainly because statistical machine learning methods produce models that are complex algebraic solutions to optimization problems such as risk minimization or geometric margin maximization. Lack of intuitive descriptions makes it hard for users to understand and verify the underlying rules that govern the model. Also, these methods cannot produce a justification for a prediction they arrive at for a new data sample. 
The problem of explaining (or justifying) a model's decision to its human user is referred to as the model interpretability problem. The sub-field is referred to as Explainable AI (XAI).

The ILP learning problem is the problem of searching for a set of logic programming clauses from which the training examples can be deduced. ILP provides an excellent solution for XAI.
ILP is a thriving field and a large number of such clause search algorithms have been devised as described by  \cite{ilp20} and \cite{ilpsurvey22}. The search in these ILP algorithms 
is performed either top down or bottom-up. A bottom-up approach builds most-specific clauses from the training examples and searches the hypothesis space by using generalization. This approach is not applicable to large-scale datasets, nor it can incorporate \textit{negation-as-failure} into the hypothesis, as explained in the book by \cite{Baral}. A survey of bottom-up ILP systems and their shortcomings has been compiled by \cite{sakama05}. In contrast, top-down approach starts with the most general clauses and then specializes them. A top-down algorithm guided by heuristics is better suited for large-scale and/or noisy datasets, as explained by \cite{quickfoil}.

The FOIL algorithm by Quinlan is a popular top-down inductive logic programming algorithm that learns a logic program. The FOLD algorithm by \cite{fold} is a novel  top-down algorithm that learns default rules along with exception(s) that closely model human thinking. It first learns default predicates that cover positive examples while avoiding covering negative examples. Then it swaps the covered positive examples and negative examples and calls itself recursively to learn the exception to the default. It repeats this process to learn exceptions to exceptions, exceptions to exceptions to exceptions, and so on. The FOLD-R++ algorithm by \cite{foldrpp} is a new scalable ILP algorithm that builds upon the FOLD algorithm to deal with the efficiency and scalability issues of the FOLD and FOIL algorithms. It introduces the prefix sum computation and other optimizations to speed up the learning process while providing human-friendly explanation for its prediction using the s(CASP) answer set programming system of \cite{scasp}. However, all these algorithms focus on binary classification tasks, and cannot deal with multi-category classification tasks. Note that a binary classification task checks whether a data record is a member of a given class or not, e.g., does a given creature fly or not fly? In multi-category classification, there can be multiple membership classes, e.g., a given creature's habitat can be predicted to be one of desert, mountain, plain, salt water, or fresh water (see the textbook by \cite{bishop2006}). 

In this paper we propose a new ILP learning algorithm called FOLD-RM for multi-category classification that builds upon the FOLD-R++ algorithm. FOLD-RM also provides native explanations for prediction without external libraries or tools. Our experimental results indicates that the FOLD-RM algorithm is comparable in performance to traditional, popular machine learning algorithms such as XGBoost by \cite{xgboost} and Multi-Layer Perceptrons (MLP) described in the book by \cite{nn-book}. 
In most cases, FOLD-RM outperforms them in execution efficiency. Of course, neither XGBoost nor MLP are interpretable. 

%GG412
Note that the term model in the field of machine learning and logic programming have  different meanings. We use the term model in this paper in machine learning sense. Thus, the answer set program generated by our FOLD-RM algorithm \textit{is the model that we learn} in the sense of machine learning. We use the term \textit{answer set} in this paper to refer to stable models of answer set programs, where a model means assignment of truth values to program predicates that make the program true. Note also that that we use the terms clause and rule interchangeably in this paper. 

\section{Background}
\label{sec:background}

\subsection{Inductive Logic Programming}
% hw 03/31
Inductive Logic Programming (ILP) as described in \cite{ilp} is a subfield of machine learning that learns models in the form of logic programming clauses comprehensible to humans. This problem is formally defined as:\\
\textbf{Given}
\begin{enumerate}
    \item A background theory $B$, in the form of an extended logic program, i.e., clauses of the form $h \leftarrow l_1, ... , l_m,\ not \ l_{m+1},...,\ not \ l_n$, where $h,l_1,...,l_n$ are positive literals and \textit{not} denotes \textit{negation-as-failure} (NAF) as described in \cite{Baral}. For reasons of efficiency, we restrict $B$ to be stratified (stratified logic programs are explained in the book by \cite{gelfondkahl}). 
    \item Two disjoint sets of ground target predicates $E^+, E^-$ known as positive and negative examples, respectively.
    \item A hypothesis language of function free predicates $L$, and a  refinement operator $\rho$ under $\theta$-subsumption described in \cite{plotkin70} (for more details see the paper by \cite{ilpsurvey22}). The hypothesis language $L$ is also assumed to be stratified.
\end{enumerate}
\textbf{Find} a set of clauses $H$ such that:
\begin{enumerate}
    \item $ \forall e \in \ E^+ ,\  B \cup H \models e$.
    \item $ \forall e \in \ E^- ,\  B \cup H \not \models e$.
    \item $B \land H$ is consistent.
\end{enumerate}

The target predicate is the predicate whose definition we want to learn as a stratified normal logic program. The positive and negative examples are grounded target predicates, i.e., suppose we want to learn the concept of which creatures can {\it fly}, then we will give positive examples $E^{+} = \{${\tt fly(tweety), fly(sam), \dots}$\}$ and negative examples $E^{-} = \{${\tt fly(kitty), fly(polly), \dots}$\}$, where {\tt tweety}, {\tt sam}, \dots, are names of creatures that can fly, and {\tt kitty}, {\tt polly}, \dots, are names of creatures that cannot fly.

Note that the reason for restricting to stratified normal logic programs is that we can realize a simple and efficient ASP interpreter in the FOLD-RM system code for the training process. If we allowed for non-stratified programs, the training process will have to invoke a full-fledged ASP interpreter during the training and testing process, resulting in significant inefficiency. Considering non-stratified programs is part of our future research plan. We restrict ourselves to function-free predicates, i.e., we allow only Datalog rules, again, for reasons of efficiency. 
%Due to the stratification requirement, all programs considered will have a unique model.

\subsection{Default Rules}

Default Logic proposed by \cite{reiter80} is a non-monotonic logic to formalize commonsense reasoning. A default $D$ is an expression of the form 

$$ A: \textbf{M} B \over\Gamma$$

\noindent which states that the conclusion $\Gamma$ can be inferred if pre-requisite $A$ holds and $B$ is justified. $\textbf{M} B$ stands for ``it is consistent to believe $B$" as explained in the book by \cite{gelfondkahl}.
Normal logic programs can encode a default quite elegantly. A default of the form: 

$$\alpha_1 \land \alpha_2\land\dots\land\alpha_n: \textbf{M} \lnot \beta_1, \textbf{M} \lnot\beta_2\dots\textbf{M}\lnot\beta_m\over \gamma$$

\noindent can be formalized as the
following normal logic program rule:

$$\gamma ~\texttt{:-}~ \alpha_1, \alpha_2, \dots, \alpha_n, \texttt{not}~ \beta_1, \texttt{not}~ \beta_2, \dots, \texttt{not}~ \beta_m.$$

\noindent where $\alpha$'s and $\beta$'s are positive predicates and \texttt{not} represents negation as failure (under the stable model semantics as described in \cite{Baral}). We call such rules default rules. 
Thus, the default $bird(X): M \lnot penguin(X)\over fly(X)$ will be represented as the following ASP-coded default rule:

~~~~~~~~~{\tt fly(X) :- bird(X), not penguin(X).}

\noindent We call {\tt bird(X)}, the condition that allows us to jump to the default conclusion that {\tt X} can fly, as the {\it default part} of the rule, and {\tt not penguin(X)} as the \textit{exception part} of the rule. 
%In the general case, these two parts will be a conjunction of predicates. 

Default rules closely represent the human thought process (i.e., frequently used in commonsense reasoning). FOLD-R and FOLD-R++ learn default rules represented as answer set programs. Note that the programs currently generated are stratified normal logic programs, however, we eventually hope to learn non-stratified answer set programs too as in the work of \cite{farhad-ilp} and \cite{shakerin-phd}. Hence, we continue to use the term answer set program for a normal logic program in this paper. An advantage of learning default rules is that we can distinguish between exceptions and noise as explained by \cite{fold} and \cite{shakerin-phd}. 
% hw 04/05
The introduction of  (nested) exceptions, or abnormal predicates, in a default rule increases coverage of the data by that default rule. A single rule can now cover more examples which results in reduced number of generated rules. The equivalent program without the abnormal predicates will have many more rules if the abnormal predicates calls are fully expanded.

\subsection{Classification Problems}

Classification problems are either binary or multi-category. 

\begin{enumerate}
    \item 
Binary classification is the task of classifying the elements of a set into two groups on the basis of a classification rule. For example, a specific patient (given a set of patients) has a particular disease or not, or a particular manufactured article (in a set of manufactured articles) will pass quality control or not. Details can be found in the book by \cite{bishop2006}. 

\item Multi-category or multinomial classification is the problem of classifying instances into one of three or more classes. For example, an animal can be predicted to have one of the following habitats: sea water, fresh water, desert, mountain, or plains. Again, details can be found in the book by \cite{bishop2006}. 
\end{enumerate}

\section{The FOLD-R++ Algorithm}
The FOLD-R++ algorithm by \cite{foldrpp} is a new ILP algorithm for binary classification that is built upon the FOLD algorithm of \cite{fold}. Our FOLD-RM algorithm builds upon the FOLD-R++ algorithm. FOLD-R++ increases the efficiency and scalability of the FOLD algorithm. The FOLD-R++ algorithms divides features into two categories: categorical features and numerical features. For a categorical feature, all the values in the feature would be considered as categorical values even though some of them are numbers. For categorical features, the FOLD-R++ algorithm only generates equality or inequality literals. For numerical features, the FOLD-R++ algorithm would try to read all the values as numbers, converting them to categorical values if conversion to numbers fails. FOLD-R++ additionally generates numerical comparison ($\leq$ and $>$) literals for numerical values. For a mixed type feature that contains both categorical values and numerical values, the FOLD-R++ algorithm treats them as numerical features.  

The FOLD-R++ algorithm employs information gain heuristic to guide literal selection during the learning process. It uses a simplified calculation process for information gain by using the number of true positive, false positive, true negative, and false negative examples that a literal can imply. The information gain for a given literal is calculated as shown in Algorithm \ref{algo:fold-r-plus-ig}.

The goal of the ILP algorithm is to find an answer set program whose answer set has all the positive examples and none of the negative examples. Our algorithm incrementally learns this program using the information gain heuristic. The Information gain heuristic allows us to refine our program incrementally, i.e., the answer set of the program after each refinement step has more and more positive examples included and fewer and fewer of the negative ones. 

\iffalse
The FOLD-R++ algorithm employs information gain heuristic to guide literal selection during the learning process. The information gain is the entropy of the given data minus the conditional entropy of the given data.
% hw 04/05
The FOLD-R++ algorithms uses a simplified calculation process for information gain by using the number of true positive, false positive, true negative, and false negative examples that a literal can imply.
% hw 04/05
The information gain for a given literal is calculated as shown in Algorithm \ref{algo:fold-r-plus-ig}. It only calculates the conditional entropy after any classification (by a new literal or a new rule) of the given data. Because the comparison of information gain of different literals only happens on the same given data, and the entropy of the given data would be exactly the same for all the literals and is unnecessary to compute.
\fi 

\begin{algorithm}[!h]
\caption{FOLD-R++ Algorithm: Information Gain function}
\label{algo:fold-r-plus-ig}
\begin{algorithmic}[1]
\Input $tp$, $fn$, $tn$, $fp$: the number of E$_{tp}$, E$_{fn}$, E$_{tn}$, E$_{fp}$ implied by literal
\Output  information gain 
\Function{F}{$a,b$}
\If{$a=0$}
\State \Return 0
\EndIf
\State \Return $a \cdot \textrm{log}_2(\frac{a}{a+b})$
\EndFunction

\Function{IG}{$tp,fn,tn,fp$} 
\If{$fp+fn>tp+tn$}
\State \Return $-\infty$
\EndIf
\State \Return $\frac{1}{tp+fp+tn+fn} \cdot $(\Call{F}{$tp,fp$} + \Call{F}{$fp,tp$} + \Call{F}{$tn,fn$} + \Call{F}{$fn,tn$})

\EndFunction
\end{algorithmic}
\end{algorithm}

% hw 03/31
The comparison between two numerical values or two categorical values in FOLD-R++ is straightforward, as commonsense would dictate, i.e., two numerical (\textit{resp.} categorical) values are equal if they are identical, else they are unequal. However, a different assumption is made to compare a numerical value and a categorical value in FOLD-R++. The equality between a numerical value and a categorical value is always false, and the inequality between a numerical value and a categorical value is always true. Additionally, numerical comparison ($\leq$ and $>$) between a numerical value and a categorical value is always false. An example is shown in Table \ref{tab:comparison} (Left), while an evaluation example for a given literal, $literal{(i,>,4)}$, based on the comparison assumption is shown in Table \ref{tab:comparison} (Right). Given E$^+=\{1,2,2,4,5,x,x,y\}$, E$^-=\{1,3,4,y,y,y,z\}$, and $literal{(i,>,4)}$, the true positive example E$_{tp}$, false negative examples E$_{fn}$, true negative examples E$_{tn}$, and false positive examples E$_{fp}$ implied by the literal are $\{5\}$, $\{1,2,2,4,x,x,y\}$, $\{1,3,4,y,y,y,z\}$, \O$ $ respectively. Then, the information gain of literal$(i,>,4)$ is calculated $IG_{(i,>,4)}(1,7,7,0)=-0.647$ through Algorithm \ref{algo:fold-r-plus-ig}.

\iffalse
\begin{table}[h]
{
\setlength{\tabcolsep}{2pt}
\begin{tabular}{|c|c|c|c|c|}
\cline{1-5}
\textbf{comparison} & 5 $=$ `k' & 5 $\neq$ `k' & 5 $\le$ `k' & 5 $>$ `k' \\
\cline{1-5}
\textbf{evaluation} & False & True & False & False \\
\cline{1-5}
\end{tabular}
}
% }
\caption{Comparisons between a numerical value and a categorical value in FOLD-R++}
\label{tab:comparison}
\end{table}
\fi

\begin{table}[]
\centering
\begin{tabular}{cc}
    \begin{minipage}{.4\linewidth}
        \centering
        \begin{tabular}{|c|c|}
        \cline{1-2}
        \textbf{comparison}  & \textbf{evaluation} \\
        \cline{1-2}
        5 $=$ `k' & False \\ 
        \cline{1-2}
        5 $\neq$ `k' & True \\ 
        \cline{1-2}
        5 $\le$ `k' & False \\ 
        \cline{1-2}
        5 $>$ `k' & False \\ 
        \cline{1-2}
        \end{tabular}
    \end{minipage} &

    \begin{minipage}{.55\linewidth}
        \centering
        \begin{tabular}{|c|c|c|}
        \cline{1-3}
          & \multicolumn{1}{c|}{\textbf{i$^{th}$ feature values}} & \textbf{count} \\
        \cline{1-3}
        $\mathbf{E^+}$ & 1 2 2 4 5 x x y & \textbf{8} \\
        \cline{1-3}
        $\mathbf{E^-}$ & 1 3 4 y y y z & \textbf{7} \\
        \cline{1-3}
        $\mathbf{E_{tp(i,>,4)}}$ & 5 & \textbf{1} \\
        \cline{1-3}
        $\mathbf{E_{fn(i,>,4)}}$ & 1 2 2 4 x x y & \textbf{7} \\
        \cline{1-3}
        $\mathbf{E_{tn(i,>,4)}}$ & 1 3 4 y y y z & \textbf{7} \\
        \cline{1-3}
        $\mathbf{E_{fp(i,>,4)}}$ & \O & \textbf{0} \\
        \cline{1-3}
        \end{tabular}
    \end{minipage} 
\end{tabular}
\caption{Left: Comparisons between a numerical value and a categorical value. Right: Evaluation and count for literal$(i,>,4)$. }
\label{tab:comparison}
\end{table}
% hw 05/03
The FOLD-R++ algorithm starts with the clause {\tt p(\ldots) :- true.}, where {\tt p(\ldots)} is the target predicate to learn. It specializes this clause by adding literals to its body during the inductive learning process. It selects a literal to add that maximizes information gain (IG). 
The literal selection process is summarized in Algorithm \ref{algo:best_ig}. In line 2, $pos$ \& $ neg$ are dictionaries that hold, respectively, the numbers of positive \& negative examples for each unique value. In line 3, $xs$ \& $cs$ are lists that hold, respectively, the unique numerical and categorical values. In line 4, $xp$ \& $xn$ are the total number of, respectively, positive \& negative examples with numerical values; $cp$ \& $cn$ are the same for categorical values. In line 11, the information gain of literal$(i,\le, x)$ is calculated by taking the parameters $pos[x]$ as the number of true positive examples, $xp-pos[x]+cp$ as the number of false negative examples, $xn-neg[x]+cn$ as the number of true negative examples, and $neg[x]$ as the number of false positive examples. After computing the prefix sum in line 6, $pos[x]$ holds the total number of positive examples that has a value less than or equal to $x$. Therefore, $xp-pos[x]$ represents the total number of positive examples that have a value greater than $x$. $cp$, the total number of positive examples that have a categorical value, is added to the number of false negative examples because of the assumption that numerical comparison between a numerical value and a categorical value is always false. The negative examples that have a value greater than $x$ or a categorical value would be evaluated as false by literal$(i,\le, x)$, so $xn-neg[x]$ is added as true negative parameter. And, $cn$, the total number of negative examples that has a categorical value, is added to true negative parameter. The expression $neg[x]$ means the number of negative examples that have the value less than or equal to $x$; $neg[x]$ is added as false positive parameter because the evaluations of these examples by literal$(i,\le, x)$ are true. The information gain calculation processes of other literals also follows the comparison assumption mentioned above. Finally, the {\tt best\_info\_gain} function returns the best score on information gain and the corresponding literal except the literals that have been used in current rule-learning process. For each feature, we compute the best literal, then the {\tt find\_best\_literal} function returns the best literal among this set of best literals.

\begin{algorithm}[!h]
\caption{FOLD-R++ Algorithm, Find Best Literal function}
\label{algo:best_ig}
\begin{algorithmic}[1]
\Input $E^+$: positive examples, $E^-$: negative examples, $L_{used}$: used literals
\Output  $best\_lit$: the best literal that provides the most information 
\Function{Best\_info\_gain}{$E^+,E^-,i,L_{used}$} 
\State $pos, neg \gets \textit{count\_classification}(E^+, E^-, i)$
% \LeftComment{pos, neg are dicts that holds the \# of pos / neg examples for each value}
\State $xs, cs \gets collect\_unique\_values(E^+, E^-, i)$ 
% \LeftComment{xs, cs are lists that holds the unique numerical and categorical values}
\State $xp, xn, cp, cn \gets count\_total(E^+, E^-, i)$  
% \LeftComment{(xp, xn) are the total \# of pos / neg examples with numerical value, (cp, cn) are the same for categorical values.}
\State $xs \gets couting\_sort(xs)$ 
\For{$j \gets 1$  to  $size(xs)$} \Comment{compute prefix sum for $E^+$ \& $E^-$ numerical values}
\State $pos[xs_{j}] \gets pos[xs_{j}] + pos[xs_{j-1}]$
\State $neg[xs_{j}] \gets neg[xs_{j}] + neg[xs_{j-1}]$
\EndFor

\For{$x \in xs$}
% \If{$literal(i,\leq, x) \in L_{used}$ or $literal(i,>, x) \in L_{used}$}
% \State continue
% \EndIf
\State $lit\_dict[literal(i,\leq, x)] \gets$ \Call{IG}{$pos[x], xp-pos[x]+cp,xn-neg[x]+cn,neg[x]$}
\State $lit\_dict[literal(i,>, x)] \gets$ \Call{IG}{$xp-pos[x],pos[x]+cp,neg[x]+cn,xn-neg[x]$}
\EndFor

\For{$c \in cs$}
% \If{$literal(i,=, x) \in L_{used}$ or $literal(i,\ne, x) \in L_{used}$}
% \State continue
% \EndIf
\State $lit\_dict[literal(i,=, x)] \gets$ \Call{IG}{$pos[c],cp-pos[c]+xp,cn-neg[c]+xn,neg[c]$}
\State $lit\_dict[literal(i,\ne, x)] \gets$ \Call{IG}{$cp-pos[c]+xp,pos[c],neg[c],cn-neg[c]+xn$}
\EndFor
\State $best, l \gets best\_pair(lit\_dict,L_{used})$
\State \Return $best, l$ \Comment{return the best info gain and its corresponding literal}
\EndFunction

\Function{Find\_best\_literal}{$E^+,E^-,L_{used}$} 
\State $best\_ig, best\_lit \gets -\infty, invalid$
\For{$i \gets 1$  to  $N$} \Comment{$N$ is the number of features}
\State $ig, lit \gets$ \Call{best\_info\_gain}{{$E^+$},{$E^-$},{i},{$L_{used}$}}
\If {$best\_ig < ig$}
\State $best\_ig, best\_lit \gets ig, lit$
\EndIf
\EndFor
\State \Return $best\_lit$
\EndFunction

\end{algorithmic}
\end{algorithm}

\begin{exmp}
\label{ex:pinguin2}
Given positive and negative examples in Table \ref{tbl:example1}, E$^+$, E$^-$, with mixed type of values on i$^{th}$ feature, the target is to find the literal with the best information gain on the given feature. There are $8$ positive examples, their values on i$^{th}$ feature are $[1,2,2,4,5,x,x,y]$, and the values on i$^{th}$ feature of the $7$ negative examples are $[1,3,4,y,y,y,z]$. 

\end{exmp}

\noindent 
With the given examples and specified feature, the number of positive examples and negative examples for each unique value are counted first, which are shown as pos, neg on right side of Table \ref{tbl:example1}. Then, the prefix sum arrays are calculated for computing heuristic as psum$^+$, psum$^-$. Table \ref{tbl:example2} shows the information gain for each literal, the literal$(i, =, x)$ has been selected with the highest score.

\begin{table}[]
\centering
\begin{tabular}{cc}
    \begin{minipage}{.4\linewidth}
        % \begin{tabular}{|c|c|}
        % \cline{1-2}
        % \textbf{examples} & ith feature values \\
        % \cline{1-2}
        % $\mathbf{E^+}$ & 1 2 2 4 5 x x y \\
        % \cline{1-2}
        % $\mathbf{E^-}$ & 1 3 4 y y y z  \\
        % \cline{1-2}
        % \end{tabular}
        \setlength{\tabcolsep}{2pt}
        \begin{tabular}{|c|c|c|c|c|c|c|c|c|}
        \cline{1-9}
          & \multicolumn{8}{c|}{\textbf{i$^{th}$ feature values}}\\
        \cline{1-9}
        $\mathbf{E^+}$ & 1 & 2 & 2 & 4 & 5 & x & x & y \\
        \cline{1-9}
        $\mathbf{E^-}$ & 1 & 3 & 4 & y & y & y & z &  \\
        \cline{1-9}
        \end{tabular}
    \end{minipage} &

    \begin{minipage}{.5\linewidth}
    \setlength{\tabcolsep}{2pt}
        \begin{tabular}{|c|c|c|c|c|c|c|c|c|}
        \cline{1-9}
        \textbf{value} & 1 & 2 & 3 & 4 & 5 & x & y & z \\
        \cline{1-9}
        \textbf{pos} & 1 & 2 & 0 & 1 & 1 & 2 & 1 & 0 \\
        \cline{1-9}
        $\mathbf{psum^+}$ & 1 & 3 & 3 & 4 & 5 & N/A & N/A & N/A \\
        \cline{1-9}
        \textbf{neg} & 1 & 0 & 1 & 1 & 0 & 0 & 3 & 1 \\
        \cline{1-9}
        $\mathbf{psum^-}$ & 1 & 1 & 2 & 3 & 3 & N/A & N/A & N/A \\
        \cline{1-9}
        \end{tabular}
    \end{minipage} 
\end{tabular}
\caption{Left: Examples and values on i$^{th}$ feature. Right: positive/negative count and prefix sum on each value }
\label{tbl:example1}
\end{table}

\begin{table}[]
\centering
{
\setlength{\tabcolsep}{2pt}
\begin{tabular}{|c|c|c|c|c|c|c|c|c|}
\cline{1-9}
 & \multicolumn{8}{c|}{\textbf{Info Gain}}  \\
\cline{1-9}
\textbf{value} & 1 & 2 & 3 & 4 & 5 & x & y & z \\
\cline{1-9}
\textbf{$\leq$ value} & $-\infty$ & -0.655 & -0.686 & -0.688 & -0.672 & N/A & N/A & N/A \\
\cline{1-9}
\textbf{$>$ value} & -0.667 & $-\infty$ & -0.682 & -0.647 & $-\infty$ & N/A & N/A & N/A \\
\cline{1-9}
\textbf{$=$ value} & N/A & N/A & N/A & N/A & N/A & -0.598 & $-\infty$ & $-\infty$  \\
\cline{1-9}
\textbf{$\ne$ value} & N/A & N/A & N/A & N/A & N/A & $-\infty$ & -0.631 & -0.637 \\
\cline{1-9}
\end{tabular}}
\caption{The info gain on i$^{th}$ feature with given examples}
\label{tbl:example2}
\end{table}

\section{The FOLD-RM Algorithm}
\label{sec:fold-rm}
%\subsection{training process}
The FOLD-R++ algorithm performs binary classification. We generalize the FOLD-R++ algorithm to perform multi-category classification. The generalized algorithm is called FOLD-RM. The FOLD-R++ algorithm is summarized in Algorithm \ref{algo:foldrpp}. The FOLD-R++ algorithm generates an answer set programming rule set, in which all the rules have the same rule head. An example covered by any rule in the set would imply the rule head is true. The FOLD-R++ algorithm generates a model by learning one rule at a time.
Ruling out the already covered example in line 9 after learning a rule would help select better literal for remaining examples. 
In the rule learning process, the best literal would be selected according to the useful information it can provide for current training examples (line 17) till the literal selection fails. If the ratio of false positive examples to true positive examples drops below the threshold $ratio$ in line 22, it would next learn exceptions by swapping residual positive and negative examples and calling itself recursively (line 26). Any examples that cannot be covered by the selected literals would be ruled out in line 20, 21. The $ratio$ in line 22 represents the upper bound on the number of true positive examples to the number of false positive examples implied by the default part of a rule. It helps speed up the training process and reduces the number of rules learned.  

Generally, avoiding covering negative examples by adding literals to the default part of a rule will reduce the number of positive examples the rule can imply. 
Explicitly activating the exception learning procedure (line 26 in Algorithm \ref{algo:foldrpp}) could increase the number of positive example a rule can cover while reducing the total number of rules generated. As a result, the interpretability is increased due to fewer rules being generated.

\begin{algorithm}[!h]
\caption{FOLD-R++ Algorithm}
\label{algo:foldrpp}
\begin{algorithmic}[1]
\Input $E^+$: positive examples, $E^-$: negative examples
\LeftComment Global Parameters: $target$, $B$: background knowledge,  $ratio$: exception ratio 
\Output  $R = \{r_1,...,r_n\}$: a set of defaults rules with exceptions 
\Function{fold\_rpp}{$E^+, E^-, L_{used}$} \Comment{$L_{used}$: used literals, initially empty}
\State $R \gets $ \O
\While{$|E^+| > 0$}
\State $r \gets$ \Call{learn\_rule}{{$E^+$}, {$E^-$}, {$L_{used}$}}
% \State $E_{TP} \gets covers(r,\ E^+,\ true)$ \Comment{$E_{TP}$: true positive examples implied by rule $r$}
\State $E_{FN} \gets covers(r,\ E^+,\ \textit{false})$ \Comment{$E_{FN}$: false negative examples implied by rule $r$}
\If {$|E_{FN}|=|E^+|$}
\State break
\EndIf
% \State $E^+ \gets E^+ \setminus E_{TP}$ \Comment{rule out the already covered examples}
\State $E^+ \gets E_{FN}$ \Comment{rule out the already covered examples}
\State $R \gets R \cup \{ r \}$
\EndWhile
\State \Return $R$
\EndFunction
\Function{learn\_rule}{${E^+}, {E^-}, {L_{used}}$}
\State $L \gets $ \O
\While{$true$}
\State  $l \gets$ \Call{find\_best\_literal}{{$E^+$}, {{$E^-$}}, {$L_{used}$}}
\State $L \gets L \cup \{ l \}$
\State $r \gets \textit{set\_default}(r,\ L)$ \Comment{set default part of rule $r$ as $L$}
\State $E^+ \gets covers(r,\ E^+,\ true)$
\State $E^- \gets covers(r,\ E^-,\ true)$
\If{$l$ is invalid or $|E^-| \leq |E^+| * ratio$}
\If{$l$ is invalid} 
\State $r \gets \textit{set\_default}(r,\ L\setminus\ \{ l \})$ \Comment{remove the invalid literal $l$ from rule $r$}
\Else   
\State $AB \gets$ \Call{fold\_rpp}{{$E^-$}, {{$E^+$}}, {$L_{used} + L$}} \Comment{learn exception rules for $r$}
\State $r \gets set\_exception(r,\ AB)$ \Comment{set exception part of rule $r$ as $AB$}
\EndIf
\State \textbf{break}
\EndIf

\EndWhile
\State \Return $r$ \Comment{the head of rule $r$ is $target$}
\EndFunction

\end{algorithmic}
\end{algorithm}

The FOLD-RM algorithm performs multi-category classification. It generates rules that it can learn for each category. If an example cannot be implied by any rule in the learned rule set, it means the model fails to classify this example. The FOLD-RM algorithm, summarized in Algorithm \ref{algo:foldrm}, first finds a target literal that represents the category with most examples among the current training set (line 4). It next splits the training set into positive and negative examples based on the target literal (line 5). Then, it learns a rule to cover the target category (line 6) by calling the {\tt learn\_rule} function of the FOLD-R++ Algorithm. The already covered examples would be ruled out from the training set in line 11, and the rule head would be changed to the target literal in line 12. However, there's a difference between the outputs of FOLD-RM and FOLD-R++. Unlike FOLD-R++, the output of FOLD-RM is a textually ordered answer set program, which means a rule is checked only if all the rules before it did not apply. The FOLD-RM system is publicly available at \url{https://github.com/hwd404/FOLD-RM}. 

Note that for learning each rule, FOLD-RM (Algorithm 4) chooses the target predicate by finding the label value with the most examples in the remaining training examples and sets it as the target predicate for this rule. In other words, the target predicate is the ``most popular" label value. The names of the predicates are the names of features in the data. The head predicate and predicates in rule body each have exactly two arguments. The first argument is a reference to the data record itself. For the target predicate, the second argument is the predicted label for that record, while for predicates in the body, the second argument is used to extract the appropriate feature value for that record. The abnormal predicates only take one argument, namely, the data record itself. 
For example, consider:
\smallskip

{\tt 

class(X,’2’) :- condition(X, ’s’), not ab5(X).

ab5(X) :- not steel(X,’r’), not enamelability(X,’2’).}

\smallskip
% hw 05/03
\noindent 
The first rule says that the predicted class of data record X is `2’ if the condition feature of X has value `s’, and abnormal case ab5 does not apply. ab5(X) is an abnormal case predicate and has only one argument. It says that the record X should not be predicted to have class value `2’, if the value of steel feature is not `r’, and the value of enamelability feature is not `2’. 

\subsection{Algorithmic Complexity}

%Theorem: Complexity of Algorithm 4 is ....
Next, we analyze the complexity of the FOLD-RM algorithm. If $M$ is the number of examples and $N$ is the number of features, it is easy to see that the time complexity of finding the best literal (Algorithm \ref{algo:best_ig}) is $O(NM)$. We assume that counting sort (complexity $O(M)$) with a pre-sorted list is used at line 5 in Algorithm \ref{algo:best_ig}. The worst case in the FOLD-RM algorithm arises when each generated rule only covers one example and each literal only excludes one non-target example. Therefore, in the worst case there will be $O(M^2)$ literals chosen in total. The worst case time complexity of the FOLD-RM algorithm (Algorithm \ref{algo:foldrm}) can be calculated to be $O(NM^3)$. However, this is a theoretical upper bound. The actual learning process is really efficient because the heuristics we employ helps select very effective literals, reducing the number of iterations in the algorithm.

One can also prove that the FOLD-RM algorithm always terminates. The {\tt fold\_rm} function calls the {\tt learn\_rule} function to induce a rule that can cover at least one `most popular' remaining example till all the examples have been covered or the learned rule fails to cover any `most popular' example. The loop in the {\tt fold\_rm} function iterates at most $|E|$ times while excluding the already covered examples. The {\tt learn\_rule}  function refines the rule with a given target by adding the best literal to the rule body. By adding literals to the rules, the numbers of true positive and false positive examples the rule implies can only monotonically decrease. The learned valid literal excludes at least one false positive example that the rule implies. So, the loop in the {\tt learn\_rule}  function iterates at most $|E^-|$ times. When the $|E^-|<|E^+|*ratio$ condition is met, the {\tt fold\_rpp} function is called to learn exception rules for the current default rule. Similar to the {\tt fold\_rm} function, the {\tt fold\_rpp} function iterates at most $|E^+|$ times. Also, there are only finite for-loops inside the {\tt find\_best\_literal} function. Therefore, we can conclude that the FOLD-RM algorithm will always terminate.

\subsection{An Illustrative Example}

We illustrate FOLD-RM, next, with a simple example. 

\begin{algorithm}[!h]
\caption{FOLD-RM Algorithm}
\label{algo:foldrm}
\begin{algorithmic}[1]
% \Input $target,B,E,ratio$ \Comment{$ratio$ is the exception ratio}
\Input $E$: examples, $B$: background knowledge,  $ratio$: exception ratio 
\Output  $R = \{r_1,...,r_n\}$: a set of defaults rules with exceptions

\Function{fold\_rm}{$E$} 
\State $R \gets $ \O
\While {$|E| > 0$}
\State $l \gets$ \Call{most}{$E$} \Comment{$l$: most popular target literal as the learning target}
\State $E^+, E^- \gets$ \Call{split\_by\_literal}{$E, l$}
\State $r \gets$ \Call{learn\_rule}{{$E^+$}, {$E^-$}, \O}
\State $E_{FN} \gets covers(r,\ E^+,\ false)$ 
\If {$|E_{FN}|=|E^+|$}
\State break
\EndIf
\State $E \gets E^+ \cup E_{FN}$ \Comment{rule out the already covered examples}
\State $r \gets add\_head(r,\ l)$
\State $R \gets R \cup \{ r \}$
\EndWhile
\State \Return $R$
\EndFunction

\Function{most}{$E$} \Comment{find the most popular target literal}
\For {$e \in E$}
\State $count[label_e] \gets count[label_e] + 1 $
\EndFor
\State $label_{most} \gets$ \Call{find\_most}{count}
\State \Return $literal(index_{label},=,label_{most})$
\EndFunction

\Function{split\_by\_literal}{$E$, $l$} 
\State $E^+, E^- \gets$ \O, \O
\For {$e \in E$}
\If {\Call{evaluate}{$e$, $l$} is true}
\State $E^+ \gets E^+ \cup \{e\}$
\Else
\State $E^- \gets E^- \cup \{e\}$
\EndIf
\EndFor
\State \Return $E^+, E^-$
\EndFunction

\end{algorithmic}
\end{algorithm}

\begin{exmp}
\label{ex:pinguin}
The target is to learn rules for \texttt{habitat} using the FOLD-RM algorithm. $ B, E$ are background knowledge and training examples, respectively. 
%GG412
There are 3 classifications: two explicit ones (land and water), and one implicit one (neither land, nor water).
\end{exmp}
\begin{verbatim}
B:  mammal(kitty).           cat(kitty).
    mammal(john).            whale(john).
    mammal(smoky).           bear(smoky).
    mammal(charlie).         dog(charlie).
    fish(nemo).              clownfish(nemo).
E:  habitat(charlie,land).   habitat(john,water).
    habitat(smoky,land).     habitat(nemo,water).
    habitat(kitty,land).
\end{verbatim}

\noindent For the first rule, the target predicate \texttt{\{habitat(X,land):- true\}} is specified at line 4 in Algorithm \ref{algo:foldrm} because `land' is the majority label. The {\tt find\_best\_literal} function selects literal \texttt{mammal(X)} as result and adds it to the clause $r$ = \texttt{\{habitat(X,land):- mammal(X)\}} at line 17 in Algorithms \ref{algo:foldrpp} because it provides the most useful information among literals \texttt{\{cat,whale,bear,dog,fish,clownfish\}}. Then the training set rules out covered examples at line 20-21 in Algorithm \ref{algo:foldrpp}, $E^+=$ \O,\ $E^-=$\texttt{\{john,nemo\}}. The default learning is finished at this point because the candidate literal cannot provide any further useful information. Therefore, the {\tt fold\_rpp} function is called recursively with swapped positive and negative examples, $E^+=$\texttt{\{john,nemo\}}, $E^-=$ \O, to learn exceptions. In this case, an abnormal predicate \texttt{\{ab1(X):-whale(X)\}} is learned and added to the previously generated clause as $r$ = \texttt{\{habitat(X,land):- mammal(X), not ab1(X)\}}. And the exception rule \texttt{\{ab1(X):- whale(X)\}} is added to the answer set program. FOLD-RM next learns rules for target predicate \texttt{\{habitat(X,water):- true\}} and two rules are generated as \texttt{\{habitat(X,water):- fish(X)\}} and \texttt{\{habitat(X,water):- whale(X)\}}. The generated final answer set program is:

\begin{center}
    \begin{tabular}{l}
        \texttt{habitat(X,land):- mammal(X), not ab1(X).} \\     
        \texttt{habitat(X,water):- fish(X).} \\
        \texttt{habitat(X,water):- whale(X).} \\
        \texttt{ab1(X):- whale(X).}
    \end{tabular}
\end{center}

\noindent The program above is a logic program, which means rules are not mutually exclusive. For correctness, a rule should be checked only if all the earlier rules result in failure. FOLD-RM generates further rules to make the learned rules mutually exclusive. The program above is transformed as shown below.

\begin{center}
    \begin{tabular}{l}
        \texttt{habitat(X,land):- habitat\_1(X).} \\
        \texttt{habitat(X,water):- habitat\_2(X), not habitat\_1(X).} \\
        \texttt{habitat(X,water):- habitat\_3(X), not habitat\_2(X), not habitat\_1(X).} \\ 
        \texttt{habitat\_1(X):- mammal(X), not ab1(X).} \\
        \texttt{habitat\_2(X):- fish(X).} \\
        \texttt{habitat\_3(X):- whale(X).} \\
        \texttt{ab1(X):- whale(X).}
    \end{tabular}
\end{center}

\section{Experimental Results}
\label{sec:Experiments}
In this section, we present our experiments on  standard UCI benchmarks. The XGBoost Classifier is a well-known classification model and used as a baseline model in our experiments. The settings used for XGBoost Classifier is kept simple without limiting its performance. Multi-Layer Perceptron (MLP) is another widely-used classification model that can deal with generic classification tasks. However, both XGBoost model and MLP cannot take mixed type (numerical and categorical values in a row or a column) as training data \textit{without pre-processing}. For mixed type data, one-hot encoding---as explained in the book by \cite{nn-book}---has been used for data preparation. For binary classification, we use accuracy, precision, recall, and F$_1$ score as evaluation metrics. For the multi-category classification tasks, following convention, we use accuracy, weighted average of Macro precision, weighted average of Macro recall, and weighted average of Macro F$_1$ score to compare models as explained by \cite{ml-multi-cat-measurement}. 
% hw 04/05 
The average numbers of generated rules are also reported for the FOLD-R++ and FOLD-RM algorithms in Table \ref{tbl:foldrppxgb}, Table \ref{tbl:foldxgb}, and Table \ref{tbl:foldmlp}, some of them are not integers because of being averaged over a number of repeated experiments.

Both FOLD-R++ and FOLD-RM algorithms \textit{do not} need any encoding for training. After specifying the numerical features, they can deal with mixed type data directly, i.e., no one-hot encoding is needed. Even missing values are handled and do not need to be provided. We implemented both algorithms with Python. The hyper-parameter $ratio$ is simply set as 0.5 for all the experiments. And all the learning processes have been run on a desktop with Intel i5-10400 CPU @ 2.9 GHz and 32 GB RAM. To have good performance test, we performed 10-fold cross-validation test on each dataset and average classification metrics and execution time are shown. The best performer is highlighted in boldface.

The XGBoost Classifier utilizes decision tree ensemble method to build model and provides good performance. Performance comparison of FOLD-R++ and XGBoost is shown in Table \ref{tbl:foldrppxgb}. The FOLD-R++ algorithm is comparable to XGBoost classifier for classification, but it's more efficient in terms of execution time, especially on datasets with many unique feature values.

\begin{table}[]
\centering
\setlength{\tabcolsep}{1pt}
\begin{tabular}{|l|c|c|c|c|c|c|c|c|c|c|c|c|c|}
% \begin{tabular}{|p{1.3cm}|p{1.3cm}|p{1.3cm}|p{.5cm}|p{.5cm}|p{.5cm}|p{.5cm}|p{.95cm}|p{.5cm}|p{.5cm}|p{.5cm}|p{.5cm}|p{.95cm}|}
\cline{1-14}
\multicolumn{3}{|c|}{DataSet} & \multicolumn{5}{c|}{XGBoost.Classifier} & \multicolumn{6}{c|}{FOLD-R++} \\ 
\cline{1-14}
Name    & \#Rows 	& \#Cols          & Acc              & Prec         & Rec          & F1            & T(ms)         & Acc          & Prec         & Rec          & F1            & T(ms)           &\#Rules\\
\cline{1-14}
acute       & 120	& 7              & \textbf{1}        & 1             & \textbf{1}    & \textbf{1}    & 35            & 0.99          & 1             & 0.99          & 0.99          & \textbf{1.6} & 2.6 \\
\cline{1-14}
autism      & 704	& 18             & \textbf{0.97}     & \textbf{0.98} & \textbf{0.98} & 0.97          & 76            & 0.95          & 0.96          & 0.97          & 0.97          & \textbf{47} & 24.3 \\
\cline{1-14}
breast-w    & 699	& 10             & 0.95              & 0.97          & 0.96          & 0.96          & 78            & \textbf{0.96} & 0.97          & 0.96          & \textbf{0.97} & \textbf{24} & 10.2\\
\cline{1-14}
cars        & 1728	& 7             & \textbf{1}        & 1             & \textbf{1}    & \textbf{1}    & 77            & 0.98          & 1             & 0.97          & 0.98          & \textbf{38} & 12.2 \\
\cline{1-14}
credit-a    & 690	& 16             & \textbf{0.85}     & 0.83          & \textbf{0.83} & 0.83          & 368           & 0.84          & \textbf{0.92} & 0.79          & \textbf{0.84} & \textbf{66} & 10.0\\
\cline{1-14}
ecoli       & 336	& 9              & 0.76              & 0.76          & 0.62          & 0.68          & 165           & \textbf{0.96} & \textbf{0.95} & \textbf{0.94} & \textbf{0.95} & \textbf{19} & 11.4 \\
\cline{1-14}
heart       & 270	& 14             & \textbf{0.80}     & \textbf{0.81} & 0.83          & 0.81          & 112           & 0.79          & 0.79          & 0.83          & 0.81          & \textbf{24} & 11.7 \\
\cline{1-14}
ionosphere  & 351 	& 35             & 0.88              & 0.86          & \textbf{0.96} & 0.90          & 1,126         & \textbf{0.92} & \textbf{0.93} & 0.94          & \textbf{0.93} & \textbf{214} & 12.0\\
\cline{1-14}
kidney      & 400 	& 25             & 0.98              & 0.98          & 0.98          & 0.98          & 126           & \textbf{0.99} & \textbf{1}    & 0.98          & \textbf{0.99} & \textbf{19} & 5.0 \\
\cline{1-14}
kr vs. kp   & 3196 	& 37            & 0.99              & 0.99          & 0.99          & 0.99          & \textbf{210}  & 0.99          & 0.99          & 0.99          & 0.99          & 223 & 18.4 \\
\cline{1-14}
mushroom    & 8124 	& 23            & 1                 & 1             & 1             & 1             & 378  & 1             & 1             & 1             & 1             & \textbf{314} & 8.0 \\
% \cline{1-14}
% sonar       & 208 	& 61             & 0.53              & 0.54          & \textbf{0.84} & 0.65          & 1,178         & \textbf{0.78} & \textbf{0.81} & 0.75          & \textbf{0.78} & \textbf{419} & 11.6 \\
\cline{1-14}
voting      & 435 	& 17             & 0.95              & 0.94          & \textbf{0.95} & 0.94          & 49            & 0.95          & 0.94          & 0.94          & 0.94          & \textbf{17} & 10.5 \\
\cline{1-14}
adult       & 32561	& 15           & \textbf{0.86}     & \textbf{0.88} & 0.94          & \textbf{0.91} & 274,655       & 0.84          & 0.86          & \textbf{0.95} & 0.90          & \textbf{2,546} & 16.7 \\
\cline{1-14}
% credit card & (30000, 24)           & -                 & -             & -             & -             & -             & \textbf{0.82} & \textbf{0.83} & \textbf{0.96} & \textbf{0.89} & \textbf{3,979} \\
% \cline{1-12}
rain in aus & 145460 & 24          & \textbf{0.83}     & 0.84          & \textbf{0.95} & \textbf{0.89} & 285,307       & 0.78          & \textbf{0.87} & 0.84          & 0.85          & \textbf{21,868} & 40.5 \\
\cline{1-14}
\end{tabular}
\caption{Comparison of XGBoost and FOLD-R++ on various Datasets}
\label{tbl:foldrppxgb}
\end{table}

\begin{table}
{
% \scriptsize
\setlength{\tabcolsep}{2pt}
\centering
\begin{tabular}{|l|c|c|l|}
\cline{1-4}
Dataset & \#Rows & \#Cols & Distribution \\
\cline{1-4}
anneal & 898 & 39 & `3': 684, `U': 40, `1': 8, `5': 67, `2': 99 \\
\cline{1-4}
avila & 20867 & 11 & `A':8572,`F':3923,`H':1039,`E':2190,`I':1663,`Y':533 \\
      &       &    & `D':705,`X':1044,`G':893,`W':89,`C':206,`B':10 \\
\cline{1-4}
ecoli & 336 & 9 & `cp':143,`im':77,`imS':2,`imL':2,`imU':35,`om':20,`omL':5,`pp':52 \\
\cline{1-4}
drug-heroin & 1885 & 13 & `CL0':1605,`CL1':68,`CL2':94,`CL3':65,`CL5':16,`CL6':13,`CL4':24 \\
\cline{1-4}
drug-crack & 1885 & 13 & `CL0':1627,`CL1':67,`CL2':112,`CL3':59,`CL5':9,`CL4':9,`CL6':2 \\
\cline{1-4}
drug-semer & 1885 & 13 & `CL0': 1877, `CL2': 3, `CL3': 2, `CL4': 1, `CL1':2 \\
\cline{1-4}
dry-bean & 13611 & 17 & `SEKER': 2027, `BARBUNYA': 1322, `BOMBAY': 522 \\
         &       &    & `CALI': 1630, `HOROZ': 1928, `SIRA': 2636, `DERMASON': 3546 \\
\cline{1-4}
eeg & 14980 & 15 & `0': 8257, `1': 6723 \\
\cline{1-4}
intention & 12330 & 18 & `FALSE': 10422, `TRUE': 1908 \\
\cline{1-4}
nursery & 12960 & 9 & `recommend': 2, `priority': 4266 \\
        &       &   & `not\_recom': 4320, `very\_recom': 328, `spec\_prior': 4044 \\
\cline{1-4}
pageblocks & 5473 & 11 & `1': 4913, `2': 329, `4': 88, `5': 115, `3': 28 \\
\cline{1-4}
parkinson & 756 & 754 & `1': 564, `0': 192 \\
\cline{1-4}
pendigits & 10992 & 17 & `8': 1055, `2': 1144, `1': 1143, `4': 1144, `6': 1056 \\
          &       &    & `0': 1143, `5': 1055, `9': 1055, `7': 1142, `3': 1055 \\
\cline{1-4}
wine & 178 & 14 & `1': 59, `2': 71, `3': 48 \\
\cline{1-4}
weight-lift & 4024 & 155 & `E': 1370, `A': 1365, `D': 276, `B': 901, `C': 112 \\
\cline{1-4}
yeast & 1484 & 10 & `MIT': 244, `NUC': 424, `CYT': 463, `ME1': 44, `EXC': 35, `ME2': 51 \\
      &      &    & `ME3': 163, `VAC': 30, `POX': 20, `ERL': 5, `0.18': 2, `0.16': 2, `0.37': 1 \\
\cline{1-4}
wall-robot & 5456 & 25 & `Slight-Right-Turn': 826, `Sharp-Right-Turn': 2097 \\
           &      &    & `Move-Forward': 2205, `Slight-Left-Turn': 328 \\
\cline{1-4}
flags & 194 & 10 & `2': 36, `6': 15, `1': 60, `0': 40, `5': 27, `3': 8, `4': 4, `7': 4 \\
\cline{1-4}
glass & 214 & 10 & `1': 70, `2': 76, `3': 17, `5': 13, `6': 9, `7': 29 \\
\cline{1-4}
optidigits & 3823 & 65 & `0': 376, `7': 387, `4': 387, `6': 377, `2': 380 \\
           &      &    & `5': 376, `8': 380, `1': 389, `9': 382, `3': 389 \\
\cline{1-4}
% wine-quality & 6497 & 12 & `6': 2836, `5': 2138, `7': 1079, `8': 193, `4': 216, `3': 30, `9': 5\\ 
% \cline{1-4}
shuttle & 58000 & 10 & {`2': 50, `4': 8903, `1': 45586, `5': 3267, `3': 171, `7': 13, `6': 10} \\ 
\cline{1-4}
\end{tabular}
\caption{The size and label distribution of UCI datasets}
\label{tbl:datasets}
}
\end{table}

For multi-category classification experiments, we collected 15 datasets for comparison with XGBoost and MLP. The drug consumption dataset has many output attributes, we perform training on heroin, crack, and semer attributes. The size and label distribution of the datasets used is shown in Table \ref{tbl:datasets}: number of rows indicates the number of data records, while the number of columns indicates the number of features.
We first compare the performance of FOLD-RM and XGBoost in Table \ref{tbl:foldxgb}. XGBoost performs much better on datasets avila and yeast, and FOLD-RM performs much better on datasets ecoli, dry-bean, eeg, and weight-lifting. After analyzing these dataset, FOLD-RM seems to perform better on more complicated datasets with mixed type values. XGBoost seems to perform better on the datasets that have limited information. However, for those datasets for which FOLD-RM has similar performance with XGBoost, FOLD-RM is more efficient in terms of execution speed. In addition, FOLD-RM is explainable/interpretable, and XGBoost is not.

\begin{table}
{
\setlength{\tabcolsep}{2pt}
% \scriptsize
\centering
\begin{tabular}{|c|c|c|c|c|c|c|c|c|c|c|c|}
% \begin{tabular}{|l|p{.55cm}|p{.55cm}|p{.55cm}|p{.55cm}|c|c|p{.55cm}|p{.55cm}|p{.55cm}|p{.55cm}|c|}
\cline{1-12}
 & \multicolumn{6}{c|}{FOLD-RM} & \multicolumn{5}{c|}{XGBoost} \\ 
\cline{1-12}
Dataset & Acc & Prec & Rec & F1 & Rules & T(ms) & Acc & Prec & Rec & F1 & T(ms) \\ 
\cline{1-12}
anneal & 0.99 & \textbf{1} & 0.99 & 0.99 & 17.9 & \textbf{63} & 0.99 & 0.99 & 0.99 & 0.99 & 295 \\
\cline{1-12}
avila & 0.33 & 0.49 & 0.33 & 0.39 & 33.6 & \textbf{3,540} & \textbf{1} & \textbf{1} & \textbf{1} & \textbf{1} & 4,897 \\
\cline{1-12}
ecoli & \textbf{0.80} & \textbf{0.82} & \textbf{0.80} & \textbf{0.80} & 42.3 & \textbf{41} & 0.42 & 0.79 & 0.42 & 0.50 & 806 \\ 
\cline{1-12}
drug-heroin & 0.84 & 0.74 & 0.84 & 0.78 & 12.0 & \textbf{136} & 0.84 & \textbf{0.77} & 0.84 & \textbf{0.79} & 1,266 \\
\cline{1-12}
drug-crack & 0.85 & 0.75 & 0.85 & 0.80 & 17.6 & \textbf{145} & 0.85 & \textbf{0.76} & 0.85 & \textbf{0.81} & 1,116 \\
\cline{1-12}
drug-semer & 0.99 & 0.99 & 0.99 & 0.99 & 10.3 & \textbf{43} & \textbf{1} & 0.99 & \textbf{1} & 0.99 & 393 \\
\cline{1-12}
dry-bean & \textbf{0.91} & \textbf{0.91} & \textbf{0.91} & \textbf{0.91} & 185.7 & 13,415 & 0.29 & 0.87 & 0.29 & 0.37 & \textbf{3,458} \\
\cline{1-12}
eeg & \textbf{0.78} & \textbf{0.78} & \textbf{0.78} & \textbf{0.77} & 164.5 & 3,914 & 0.50 & 0.70 & 0.50 & 0.54 & \textbf{340} \\
\cline{1-12}
intention & 0.90 & 0.89 & 0.90 & \textbf{0.90} & 78.3 & \textbf{1,621} & 0.90 & 0.89 & 0.90 & 0.89 & 114,161 \\
\cline{1-12}
nursery & \textbf{0.97} & \textbf{0.97} & \textbf{0.97} & \textbf{0.96} & 59.8 & \textbf{643} & 0.88 & 0.93 & 0.88 & 0.89 & 24,100 \\ 
\cline{1-12}
pageblocks & \textbf{0.97} & \textbf{0.97} & \textbf{0.97} & \textbf{0.96} & 72.3 & \textbf{929} & 0.95 & 0.94 & 0.95 & 0.94 & 81,416 \\ 
\cline{1-12}
parkinson & 0.81 & 0.80 & 0.81 & 0.79 & 15.9 & 8,503 & \textbf{0.84} & \textbf{0.84} & \textbf{0.84} & \textbf{0.83} & \textbf{527} \\
\cline{1-12}
pendigits & \textbf{0.96} & \textbf{0.96} & \textbf{0.96} & \textbf{0.96} & 219.2 & \textbf{2,447} & 0.91 & 0.92 & 0.91 & 0.91 & 54,102 \\
\cline{1-12}
wine & \textbf{0.94} & 0.97 & \textbf{0.94} & 0.95 & 7.6 & \textbf{17} & 0.93 & \textbf{1} & 0.93 & \textbf{0.96} & 49 \\
\cline{1-12}
weight-lift & \textbf{1} & \textbf{1} & \textbf{1} & \textbf{1} & 14.0 & \textbf{1,879} & 0.51 & 0.81 & 0.51 & 0.57 & 224,140 \\
\cline{1-12}
yeast & 0.08 & 0.15 & 0.08 & 0.10 & 8.7 & \textbf{146} & \textbf{0.45} & \textbf{0.5} & \textbf{0.45} & \textbf{0.45} & 8,629 \\
\cline{1-12}
wall-robot & 0.99 & 0.99 & 0.99 & 0.99 & 30.5 & 2,402 & 0.99 & 0.99 & 0.99 & 0.99 & \textbf{403} \\
\cline{1-12}
\end{tabular}
\caption{Comparison of FOLD-RM and XGBoost on UCI Datasets}
\label{tbl:foldxgb}
}
\end{table}

\begin{table}
{
% \scriptsize
\setlength{\tabcolsep}{2pt}
\centering
\begin{tabular}{|c|c|c|c|c|c|c|c|c|c|c|c|}
% \begin{tabular}{|l|p{.55cm}|p{.55cm}|p{.55cm}|p{.55cm}|c|c|p{.55cm}|p{.55cm}|p{.55cm}|p{.55cm}|c|}
\cline{1-12}
 & \multicolumn{6}{c|}{FOLD-RM} & \multicolumn{5}{c|}{MLP} \\ 
\cline{1-12}
% Dataset & Acc. & Prec. & Rec. & F1 & rules & T(ms) & Acc. & Prec. & Rec. & F1 & T(ms) \\ 
Dataset & Acc & Prec & Rec & F1 & Rules & T(ms) & Acc & Prec & Rec & F1 & T(ms) \\ 
\cline{1-12}
anneal & 0.99 & \textbf{1} & 0.99 & 0.99 & 17.9 & \textbf{63} & 0.99 & 0.99 & 0.99 & 0.99 & 462 \\
\cline{1-12}
avila & 0.33 & 0.49 & 0.33 & 0.39 & 33.6 & \textbf{3,540} & \textbf{0.90} & \textbf{0.90} & \textbf{0.90} & \textbf{0.90} & 73,610 \\
\cline{1-12}
ecoli & \textbf{0.80} & 0.82 & \textbf{0.80} & \textbf{0.80} & 42.3 & \textbf{41} & 0.52 & \textbf{0.91} & 0.52 & 0.61 & 411 \\ 
\cline{1-12}
drug-heroin & \textbf{0.84} & 0.74 & \textbf{0.84} & 0.78 & 12.0 & \textbf{136} & 0.82 & \textbf{0.77} & 0.82 & \textbf{0.79} & 1,093 \\
\cline{1-12}
drug-crack & \textbf{0.85} & 0.75 & \textbf{0.85} & 0.80 & 17.6 & \textbf{145} & 0.84 & 0.77 & 0.84 & 0.80 & 1,061 \\
\cline{1-12}
drug-semer & 0.99 & 0.99 & 0.99 & 0.99 & 10.3 & \textbf{43} & \textbf{1} & 0.99 & \textbf{1} & 0.99 & 518 \\
\cline{1-12}
dry-bean & \textbf{0.91} & 0.91 & \textbf{0.91} & \textbf{0.91} & 185.7 & 13,415 & 0.57 & \textbf{0.92} & 0.57 & 0.66 & \textbf{11,292} \\
\cline{1-12}
eeg & \textbf{0.78} & \textbf{0.78} & \textbf{0.78} & \textbf{0.77} & 164.5 & \textbf{3,914} & 0.49 & 0.68 & 0.49 & 0.54 & 5,946 \\
\cline{1-12}
intention & \textbf{0.90} & \textbf{0.89} & \textbf{0.90} & \textbf{0.90} & 78.3 & \textbf{1,621} & 0.84 & 0.76 & 0.84 & 0.78 & 218,087 \\
\cline{1-12}
nursery & \textbf{0.97} & \textbf{0.97} & \textbf{0.97} & \textbf{0.96} & 59.8 & \textbf{643} & 0.91 & 0.94 & 0.91 & 0.91 & 943 \\ 
\cline{1-12}
pageblocks & \textbf{0.97} & \textbf{0.97} & \textbf{0.97} & \textbf{0.96} & 72.3 & \textbf{929} & 0.93 & 0.91 & 0.93 & 0.92 & 6,452 \\ 
\cline{1-12}
parkinson & 0.81 & 0.80 & 0.81 & 0.79 & 15.9 & 8,503 & \textbf{0.82} & \textbf{0.82} & \textbf{0.82} & \textbf{0.81} & \textbf{1,416} \\
\cline{1-12}
pendigits & 0.96 & 0.96 & 0.96 & 0.96 & 219.2 & \textbf{2,447} & \textbf{0.99} & \textbf{0.99} & \textbf{0.99} & \textbf{0.99} & 6,732 \\
\cline{1-12}
wine & 0.94 & 0.97 & 0.94 & 0.95 & 7.6 & \textbf{17} & \textbf{0.97} & \textbf{1} & \textbf{0.97} & \textbf{0.98} & 189 \\
\cline{1-12}
weight-lift & \textbf{1} & \textbf{1} & \textbf{1} & \textbf{1} & 14.0 & \textbf{1,879} & 0.54 & 0.89 & 0.54 & 0.58 & 52,643 \\
\cline{1-12}
yeast & 0.08 & 0.15 & 0.08 & 0.10 & 8.7 & \textbf{146} & \textbf{0.41} & \textbf{0.49} & \textbf{0.41} & \textbf{0.38} & 3,750 \\
\cline{1-12}
wall-robot & \textbf{0.99} & \textbf{0.99} & \textbf{0.99} & \textbf{0.99} & 30.5 & \textbf{2,402} & 0.88 & 0.88 & 0.88 & 0.88 & 8,141 \\
\cline{1-12}
\end{tabular}
\caption{Comparison of FOLD-RM and MLP on UCI Datasets}
\label{tbl:foldmlp}
}
\end{table}

The comparison with MLP is presented in Table \ref{tbl:foldmlp}. For most datasets, FOLD-RM can achieve equivalent scores, similar to the comparison with XGBoost, FOLD-RM performs much better on datasets ecoli, dry-bean, eeg, and weight-lifting, while MLP performs much better on datasets avila and yeast. MLP takes much more time for training than XGBoost because of its algorithmic complexity. 
Like the XGBoost classifier, for complex datasets with mixed values, MLP also suffers from pre-processing complications such as having to use one-hot encoding.  

\iffalse
\begin{table}[h]
{
\setlength{\tabcolsep}{2pt}
\begin{tabular}{|l|c|c|}
\cline{1-3}
Dataset & RIPPER Acc & FOLD-RM Acc \\
\cline{1-3}
ecoli & 0.80 & 0.80 \\
\cline{1-3}
flags & \textbf{0.61} & 0.58 \\
\cline{1-3}
glass & 0.63 & 0.63 \\
\cline{1-3}
nursery & 0.72 & \textbf{0.96} \\
\cline{1-3}
optidigits & 0.90 & 0.90 \\ 
\cline{1-3}
pageblocks & \textbf{0.97} & 0.96 \\
\cline{1-3}
pendigits & 0.95 & 0.95 \\
\cline{1-3}
wine-quality & 0.54 & \textbf{0.58} \\
\cline{1-3}
\end{tabular}
}
% }
\caption{Comparison of RIPPER and FOLD-RM on UCI Datasets}
\label{tab:ripperfoldrm}
\end{table}
\fi 

\begin{table}[h]
{
\setlength{\tabcolsep}{2pt}
\begin{tabular}{|l|c|c||l|c|c|}
\cline{1-6}
Dataset & RIPPER Acc & FOLD-RM Acc & Dataset & RIPPER Acc & FOLD-RM Acc\\
\cline{1-6}
ecoli & 0.80 & 0.80 & flags & \textbf{0.61} & 0.58 \\
\cline{1-6}
glass & 0.63 & 0.63 & nursery & 0.72 & \textbf{0.96} \\
\cline{1-6}
optidigits & 0.90 & 0.90 & pageblocks & \textbf{0.97} & 0.96\\ 
\cline{1-6}
pendigits & 0.95 & 0.95 & shuttle & 0.99 & \textbf{1} \\
\cline{1-6}
\end{tabular}
}
% }
\caption{Comparison of RIPPER and FOLD-RM on UCI Datasets}
\label{tab:ripperfoldrm}
\end{table}

RIPPER algorithm by \cite{RIPPER} is a popular rule induction algorithm that generates conjunctive normal form (CNF) formulas. Eight datasets have been used for comparison between RIPPER and FOLD-RM. We did not find the RIPPER algorithm implementation with multi class classification. Therefore, we have collected the accuracy data reported by \cite{ripmc} and performed the same experiment with the same datasets with the FOLD-RM algorithm. Two-thirds of the dataset was used for training by \cite{ripmc} and the remaining one-third used as the test set. We follow the same convention. For each dataset, this process was repeated 50 times. The average of accuracy is shown in Table \ref{tab:ripperfoldrm}. Both algorithms have similar accuracy on most datasets, though FOLD-RM outperforms on nursery dataset. Ripper is explainable, as it outputs CNF formulas. However, the CNF formulae generated tend to have large number of literals. In contrast, FOLD-RM rules are succinct due to use of negation as failure and they have an operational semantics (that aligns with how humans reason) by virtue of being a normal logic program.

\section{Prediction and Justification}

The FOLD-RM algorithm generates rules that can be interpreted by the human user to understand the patterns and correlations that are implicit in the table data. These rules can also be used to make prediction given new data input. Thus FOLD-RM serves as a machine learning algorithm in its own right.
However, making good predictions is not enough for  critical tasks such as disease diagnosis and loan approval. 
%The prediction needs to be justified as well. FOLD-R++ uses the s(CASP) answer set programming system \cite{scasp} to provide justification for a prediction. The answer set program generated is loaded into the s(CASP) system, the feature values of the new case (for which prediction has to be made) are coded as facts, and the query issued to compute the prediction \cite{foldrpp}. The justification facility of s(CASP) can be used to generate a justification \cite{scasp-just}. Prediction and justification can be provided for FOLD-RM learned rules in a similar manner. 
FOLD-RM comes with a built-in prediction and justification facility. We illustrate this justification facility via an example.

\begin{exmp}
\label{ex:anneal}
The ``annealing" UCI dataset is a multi-category classification task which contains 798 training examples and 100 test examples and their classes based on features such as steel, carbon, hardness, condition, strength, etc. FOLD-RM generates the following answer set program with 20 rules for 5 classes, which is pretty concise and precise:
\end{exmp}
{
\scriptsize
\begin{verbatim}
        classes(X,'3') :- not surface_quality(X,'?'), not ab1(X), 
                        not ab2(X), not ab3(X), not ab4(X). 
        classes(X,'2') :- condition(X,'s'), not ab5(X). 
        classes(X,'3') :- not carbon(X,'00'), not ab6(X). 
        classes(X,'5') :- family(X,'tn'). 
        classes(X,'u') :- steel(X,'a'), not ab7(X). 
        classes(X,'2') :- thick(X,N32), N32>0.8, not ab8(X), 
                      not ab9(X), not ab10(X). 
        classes(X,'3') :- not steel(X,'s'), not ab11(X), not ab6(X). 
        classes(X,'1') :- family(X,'?'). 
        classes(X,'1') :- family(X,'zs'). 
        ab1(X) :- hardness(X,'85'). 
        ab2(X) :- strength(X,'600').               ab3(X) :- carbon(X,'10'). 
        ab4(X) :- hardness(X,'80'), cbond(X,'?'). 
        ab5(X) :- not steel(X,'r'), not enamelability(X,'2'). 
        ab6(X) :- steel(X,'a').                    ab7(X) :- carbon(X,'03'). 
        ab8(X) :- steel(X,'r').                    ab9(X) :- steel(X,'s'). 
        ab10(X) :- not temper_rolling(X,'?').      ab11(X) :- not family(X,'?'). 
\end{verbatim}}

The above generated rule set achieves 0.99 accuracy, 0.99 weighted Macro precision, 0.99 weighted Macro recall, and 0.99 weighed Macro F1 score. The justification tree generated by the FOLD-RM system for the 8$^{th}$ test example is shown below:

{
\scriptsize
\begin{verbatim}
    Proof Tree for example number 8 :
    the value of classes is 2 DOES HOLD because 
        the value of condition is 's' which should equal 's' (DOES HOLD) 
        exception ab5 DOES NOT HOLD because 
            the value of steel is 'r' which should not equal 'r' (DOES NOT HOLD) 
            the value of enamelability is '?' which should not equal '2' (DOES HOLD) 
    {'condition: S', 'enamelability: ?', 'steel: R'}
\end{verbatim}  
}

\noindent This justification tree is also shown in another format: by showing which rules were involved in the proof/justification. For each call in each rule that was invoked, FOLD-RM shows whether it is true ([T]) or false ([F]). The head of each applicable rule is similarly annotated. 
%The proof tree in both format are relatively straightforward to generate, and so details of the generation process are omitted.
We illustrate this for the 8$^{th}$ test example:
{
\scriptsize
\begin{verbatim}
    [F]ab5(X) :- not [T]steel(X,'r'), not [F]enamelability(X,'2'). 
    [T]classes(X,'2') :- [T]condition(X,'s'), not [F]ab5(X). 
    {'condition: S', 'enamelability: ?', 'steel: R'}
    \end{verbatim}  
}

\section{Conclusions and Related Work}
\label{sec:concl}
In this paper we presented FOLD-RM, an efficient and highly scalable algorithm for multi-category classification tasks. FOLD-RM can generate explainable answer set programs and human-friendly justification for predictions. Our algorithm does not need any encoding (such as one-hot encoding) for data preparation. Compared to the well-known classification models like XGBoost and MLP, our new algorithm has similar performance in terms of accuracy, weighted macro precision, weighted macro recall, and weighted macro F$_1$ score. However, our new approach is much more efficient and interpretable than these other approaches. It is remarkable that an ILP system is comparable in accuracy to state-of-the-art traditional machine learning systems. 

ALEPH by \cite{aleph} is a well-known ILP algorithm that employs bottom-up approach to induce rules for non-numerical data. Also, no automatic method is available for the specialization process. A tree-ensemble based rule extraction algorithm is proposed by  \cite{Akihiro2021}, its performance relies on trained tree-ensemble model. It may also suffer from scalability issue because its running time is exponential in the number of valid rules. 

In practice, statistical Machine Learning models show good performance for classification. Extracting rules from statistical models is also a long-standing research topic.  Rule extraction algorithms are of two kinds: 1) Pedagogical (learning rules from black box models without looking into internal structures), such as, TREPAN by \cite{trepan}, which learns decision trees from neural networks 2) Decompositional (learning rules by analysing the models inside out) such as, SVM+Prototypes by \cite{proto-svm}, which employs clustering algorithm to extract rules from SVM classifiers by utilizing support vectors. RuleFit by Friedman and Popescu is another rule extraction algorithm that learns sparse linear models with original feature decision rules from shallow tree ensemble model for both classification and regression tasks. However, its interpretability decreases when too many decision rules have been generated. Also, simpler approaches that are a combination of statistical method with ILP have been extensively explored. The kFOIL system by \cite{kfoil} incrementally learns kernel for SVM FOIL style rule induction. The nFOIL system by \cite{nfoil} is an integration of Naive Bayes model and FOIL. TILDE by  \cite{tilde} is another top-down rule induction algorithm based on C4.5 decision tree, it can achieve similar performance with Decision Tree. However, it would suffer from scalability issue when there are too many unique numerical values in the dataset. For most datasets we experimented with, the number of leaf nodes in the trained C4.5 decision tree is much more than the number of rules that FOLD-R++/FOLD-RM generate. The FOLD-RM algorithm outperforms
the above methods in efficiency and scalability due to (i) its use of learning defaults, exceptions to defaults, exceptions to exceptions, and so on (i) its top-down nature, and (iii) its use of improved method (prefix sum) for heuristic calculation. 

%Our future work includes extending FOLD-R++ and FOLD-RM to learn non-stratified answer set programs.

\section*{Acknowledgment}
Authors acknowledge support from NSF grants IIS 1718945, IIS 1910131, IIP 1916206, US DoD, Atos Corp and Amazon Corp. We thank our colleagues Joaquin Arias, Parth Padalkar, Kinjal Basu, Sarat Chandra Varanasi, Elmer Salzar, Fang Li, Serdar Erbatur, and Doug DeGroot for discussions and help.

\bibliographystyle{tlplike}
\bibliography{mycitations}

%\label{lastpage}
\end{document}